\documentclass[preprint,5p]{elsarticle} 

\makeatletter
\def\ps@pprintTitle{%
 \let\@oddhead\@empty
 \let\@evenhead\@empty
 \let\@oddfoot\@empty
 \let\@evenfoot\@empty
}
\makeatother

\usepackage{amssymb}
\usepackage{amsmath}

\journal{Solar Energy}

\usepackage{siunitx}
\usepackage{mathtools}
\usepackage{hyperref}
\usepackage{cleveref}
\usepackage{url}
\usepackage{booktabs}
\usepackage{multirow}
\usepackage{multicol}

\DeclareMathOperator{\sinc}{sinc}

\newtheorem{theorem}{Theorem}[section]

\newtheorem{observation}[theorem]{Observation}

\newcommand{\ET}{E} 
\newcommand{\E}{dE}

\newcommand{\ED}{E_D} 
\newcommand{\EB}{E_B} 

\newcommand{\Dth}{\Delta\theta}
\newcommand{\nv}{\vec{n}}
\newcommand{\sv}{\vec{s}}
\newcommand{\dw}{d\omega} 
\newcommand{\Ls}{L(\sv)} 
\newcommand{\Ln}{L(\nv)} 
\newcommand{\ETn}{\ET(\nv)}
\newcommand{\Rs}{R_{\sv}} 
\newcommand{\hz}{\hat{z}} 
\newcommand{\kn}{k(\nv)} 

\newcommand{\Lth}{L(\theta)}
\newcommand{\Esn}{\E(\sv;\,\nv)} 
\newcommand{\ETth}{\ET(\theta)} 
\newcommand{\EDth}{\ED(\theta;\Dth)} 
\newcommand{\Hjw}{H(j\omega)} 
\newcommand{\kth}{k(\theta)} 
\newcommand{\bth}{b(\theta;\Dth)} 
\newcommand{\EBth}{\EB(\theta;\Dth)} 

\newcommand{\Kjw}{K(j\omega)}
\newcommand{\Bjw}{B(j\omega)}

\newcommand{\urbanskyweb}[1]{\href{https://cave.cs.columbia.edu/repository/UrbanSky}{#1}}
\newcommand{\projectweb}[1]{\href{https://cave.cs.columbia.edu/projects/categories/project?cid=Computational\%20Imaging&pid=Minimal\%20Sensing\%20for\%20Orienting\%20a\%20Solar\%20Panel}{#1}}

\begin{document}

\begin{frontmatter}



\title{Minimal Sensing for Orienting a Solar Panel}


\author[1]{Jeremy Klotz\corref{cor1}} 
\ead{jklotz@cs.columbia.edu}
\cortext[cor1]{Corresponding author}

\author[1]{Shree K. Nayar} 
\ead{nayar@cs.columbia.edu}

\affiliation[1]{organization={Computer Science Department, Columbia University},
            addressline={500 West 120th St.}, 
            city={New York},
            postcode={10027}, 
            state={NY},
            country={USA}}

\begin{abstract}

A solar panel harvests the most energy when pointing in the direction that maximizes the total illumination (irradiance) falling on it. Given an arbitrary panel orientation and an arbitrary environmental illumination, we address the problem of finding the direction of maximum total irradiance. We develop a minimal sensing approach where measurements from just four photodetectors are used to iteratively vary the tilt of the panel to maximize the irradiance. Many environments produce irradiance functions with multiple local maxima. As a result, simply measuring the gradient of the irradiance function and applying gradient ascent will not work. We show that a larger, optimized tilt between the detectors and the panel is equivalent to blurring the irradiance function. This has the effect of eliminating local maxima and turning the irradiance function into a unimodal one, whose maximum can be found using gradient ascent. We show that there is a close relationship between our approach and scale space theory. We collected a large dataset of high-dynamic range lighting environments in Manhattan, called \textit{UrbanSky}. We use this dataset to conduct simulations to verify the robustness of our approach. Next, we simulate the energy harvested using our approach under dynamic illumination. Finally, we built a portable solar panel with four compact detectors and an actuator to conduct experiments in various real-world settings: direct sunlight, cloudy sky, urban settings with occlusions and shadows, and complex indoor lighting. In all cases, we show improvements in harvested energy compared to standard approaches for orienting a solar panel.

\end{abstract}


\end{frontmatter}


\section{The Orientation of a Solar Panel}
Solar panels are widely deployed in open fields, on rooftops, and in urban settings to harvest energy from sunlight. Consider a panel in a desert on a sunny day. Since the sun is the dominant light source, the illumination from the entire sky can be approximated as coming from a single point (the sun). Thus, pointing the panel in the direction of the sun would maximize the irradiance\footnote{Unless otherwise specified, by irradiance we mean the total irradiance received by the panel from the entire hemisphere visible to it.} of the panel and hence the energy harvested by it.\footnote{Due to non-idealities in the energy harvesting system, the harvested energy from a solar panel may be a non-linear function of its irradiance. However, since this function is monotonic, the orientation that maximizes the irradiance also maximizes the harvested energy.} In this case, the panel can track the sun as its trajectory is easily determined from the latitude and longitude of the panel, the date, and the time of day~\cite{michalskyAstronomicalAlmanacAlgorithm1988, meeusAstronomicalAlgorithms1991}. 

But what is the optimal solar panel orientation on a cloudy day? Now, the sky is an extended light source with a possibly complex radiance function that includes multiple peaks in different directions. In this setting, since a solar panel aggregates light from an entire hemisphere, simply orienting it towards the brightest point in the sky is almost certain to not maximize the irradiance. 

\begin{figure}[t!]
    \centering
    \includegraphics[]{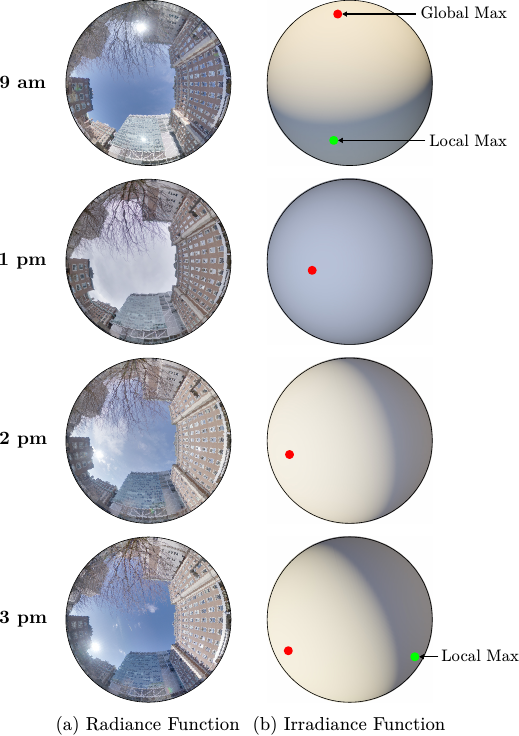}
    \caption{
    \textbf{Complex Illumination in Urban Environments.}
    (a)~The radiance function of an urban environment, which describes the environmental illumination, at different times of day. As the sun moves through the sky, the illumination varies dramatically due to changing weather, shadows, and reflections of the sun.
    (b)~The irradiance function, which specifies the total irradiance of a solar panel as a function of the panel orientation, varies significantly with the environmental illumination. In urban settings such as this one with complex illumination, the irradiance function often has multiple local maxima (green and red dots). As a result, iteratively tilting a solar panel based on the gradient of the irradiance function would produce sub-optimal orientations. Given any environment, our goal is to orient a solar panel in the direction of maximum irradiance (red dot) using minimal sensing and processing, regardless of the complexity of the illumination.
    }
    \label{fig:intro}
    \vspace{-0.1in}
\end{figure}

The problem gets even more interesting in urban settings. Consider a solar panel in the urban environment shown in \cref{fig:intro}(a). 
In this case, the panel only sees the sun for a fraction of the day due to occlusions by nearby buildings. When the sun is occluded, the panel would only be illuminated by a patch of the sky and reflections from buildings and objects around it. As can be seen in \cref{fig:intro}(a), even within a single day, the lighting may change dramatically due to shadows and reflections of the sun. Many cities have deployed large numbers of solar panels in dense urban environments to power sensors, lamps, and devices such as electric bikes. Currently, these panels are either fixed in orientation or simply track the sun.

Beyond dense urban environments, solar panels are increasingly used indoors to harvest energy from both the indoor and outdoor illumination to power environmental sensors and consumer devices. In this scenario, the illumination can be expected to vary significantly, causing the direction of maximum irradiance to shift dramatically throughout the day. Numerous smart city applications stand to benefit from indoor solar panels that can continuously track the direction of maximum irradiance. The market for indoor panels is growing rapidly and is projected to be \$154 million by 2030~\cite{precisionreportsGlobalIndoorSolar2024}. 

The goal of our work is to find the direction of maximum irradiance in {\em any environment} using minimal sensing resources. The first instinct may be to use a fisheye camera placed in the environment to measure the incoming radiance from every direction and compute the direction of maximum irradiance. Traditional cameras, however, not only add cost to the system but are also power-hungry---the image sensor alone consumes hundreds of milliwatts~\cite{likamwaEnergyCharacterizationOptimization2013}, which, in effect, reduces the energy harvested by the panel. While this cost and energy overhead can be ignored in the case of large arrays of panels, it would be prohibitive in the case of smaller stand-alone panels. {\em For applications involving a single small panel, a minimal sensing approach that captures the fewest light measurements and requires negligible processing is highly desirable.}

The goal of this work is not to propose a product, as there are existing products in the market that involve a single solar panel mounted on an actuator. Instead, our goal is to analyze a minimal sensing approach for orienting a solar panel in any lighting environment. Since our approach uses minimal sensing and processing, it can be directly applied to existing products with little additional complexity. 

One approach would be to measure the gradient of the incident irradiance along each of the two dimensions of the panel using four photodetectors that are slightly tilted with respect to each other. The measurements can then be used to perform gradient ascent to iteratively tilt the panel towards the direction of maximum irradiance. Unfortunately, the irradiance function of a complex lighting environment, such the urban environment in \cref{fig:intro}(a), is likely to have multiple local maxima, as shown in \cref{fig:intro}(b). As a result, such a naive approach to orient a solar panel would produce sub-optimal panel orientations. 

This brings us to our key result. Using Fourier analysis, we show that by using carefully chosen larger tilts between the four photodetectors, we can find the gradient of a function that is the original irradiance function convolved with a box filter of a specific width. Even when the original irradiance function has multiple modes, the convolved (blurred) function is almost certain to have a single mode that is close to the largest mode of the original function. This result is closely related to scale space theory~\cite{witkinScaleSpaceFiltering1987, lindebergScaleSpaceTheoryComputer1994}. Therefore, by simply computing finite differences between the measurements produced by four carefully oriented detectors and using them to perform gradient ascent, the panel converges to the orientation that yields maximum irradiance. 

To validate our approach in simulation, we have collected a large dataset of high-dynamic range (HDR) lighting environments in New York City. This dataset, called \textit{UrbanSky}, consists of 1,067 lighting environments under various weather conditions and at different times of day. In each environment in \textit{UrbanSky}, we capture the illumination using a $360\unit{\degree}$ camera, we measure the global horizontal irradiance using a pyranometer, and we record the date, time of day, GPS location, and current weather conditions.
Using \textit{UrbanSky}, we simulate our approach for orienting a solar panel to identify the range of detector tilt angles that yield the best panel orientation. Next, we use \textit{UrbanSky} to compare the energy harvested using our approach with alternative strategies that use minimal sensing to orient a solar panel. Across the 1,067 scenes in \textit{UrbanSky}, orienting a solar panel using our approach increases the harvested energy. We have released the code to benchmark our approach against alternative strategies \projectweb{online}. In addition, the \textit{UrbanSky} dataset is \urbanskyweb{available online} to encourage future work both in solar energy and computer graphics.

We conduct simulations to evaluate the energy harvested using our approach under dynamic illumination. To this end, we used a physically based renderer and a 3D model of a city to simulate the illumination seen by a solar panel across 6,000 different locations in an urban environment. We compare the energy harvested using our approach with the energy harvested using two common strategies: (a)~fixing the orientation of the panel, and (b)~tracking the sun. Across the 6,000 solar panel locations, the panel oriented using our approach harvests more energy over an entire day.

We test our approach in real-world settings using a prototype system. Our prototype includes a single solar panel with four compact detectors attached to its periphery, and the panel is mounted on a two-axis actuator. 
We compare the energy harvested by our prototype with the energy harvested by a solar panel in a fixed orientation and another that tracks the sun. In diverse real-world settings---direct sunlight, cloudy sky, urban settings with occlusions, shadows, and reflections, and an indoor room with complex lighting---our approach increases the harvested energy.

\section{Related Work}
Solar panels are either fixed at a specific orientation or mounted on actuators that vary their orientation to increase the harvested energy. With respect to fixed panel orientations~\cite{duffieSolarEngineeringThermal2006, gunerhanDeterminationOptimumTilt2007, laveOptimumFixedOrientations2011, lubitzEffectManualTilt2011, jacobsonWorldEstimatesPV2018}, a widely used rule of thumb is to orient the panel towards the equator with an angle from the zenith equal to the panel's latitude~\cite{duffieSolarEngineeringThermal2006}. Panels that are mounted on actuators to move throughout the day almost exclusively track the sun~\cite{mousazadehReviewPrincipleSuntracking2009, leeSunTrackingSystems2009, al-rousanAdvancesSolarPhotovoltaic2018, awasthiReviewSunTracking2020}. Kelly and Gibson~\cite{kellyImprovedPhotovoltaicEnergy2009} observed that tracking the sun is sub-optimal on a cloudy day and suggested pointing the panel straight up at the sky when the sun is occluded by clouds. All of these prior works only consider outdoor environments with an unobstructed view of the sky. Our work considers a more general problem: we wish to find the panel orientation that yields maximum irradiance in an arbitrary environment. This problem has become highly relevant as panels are now being used in environments that include occlusions, shadows, and multiple dominant sources. We demonstrate that our approach increases the harvested energy in complex lighting environments compared to a panel that is fixed in orientation~\cite{duffieSolarEngineeringThermal2006} or one that tracks the sun.

It is widely accepted that solar energy harvesting in urban environments, in particular, is an increasingly important application of photovoltaics. Orienting a solar panel in urban environments is a complex problem since one must consider not only the illumination from the sky but also reflections and shadows from nearby infrastructure. Prior work seeks to find the best \textit{fixed} orientation in urban settings by using knowledge of the scene's 3D structure and material properties to compute the irradiance falling on rooftops and building façades~\cite{mardaljevicIrradiationMappingComplex2003, robinsonSolarRadiationModelling2004, compagnonSolarDaylightAvailability2004, jakubiecMethodPredictingCitywide2013, jakicaStateoftheartReviewSolar2018, ProjectSunroof}. All of these works account for the complex illumination caused by city infrastructure. The goal of our work is different: given a solar panel in any lighting environment, which may include shadows and reflections in urban settings, we use an actuator to iteratively tilt the panel toward the direction of maximum irradiance using minimal sensor measurements. To encourage future work in the analysis of urban lighting environments, we have released \textit{UrbanSky}, a dataset of 1,067 outdoor HDR lighting environments captured in New York City. 

A variety of methods seek to find the optimal orientation of a solar panel by iteratively tilting the panel based on measurements produced by light sensors. Most closely related to our work are methods that use detectors mounted on the panel at an angle to produce differential measurements~\cite{lynchSimpleElectroOpticallyControlled1990, poulekNewSolarTracker1998, poulekVerySimpleSolar2000, sefaApplicationOneaxisSun2009, awayDualaxisSunTracker2017}. When the panel points toward the sun on a clear day, the detectors are illuminated equally, and the differential measurement is zero. Thus, the tracking algorithm seeks to minimize the differential measurement in order to track the sun. 
While this approach can orient a panel toward the sun on a clear day, its convergence in arbitrary lighting is not guaranteed. This is because, in all the previous work, the tilt angle between the detectors and the solar panel is chosen in an ad-hoc manner. 
We provide a detailed analysis of the differential measurements produced by tilted detectors. We first show that the measurement is equal to the derivative of a function that results from blurring the irradiance function with a box filter. When the detector tilt angle is carefully chosen, the blurred irradiance function is unimodal, regardless of the complexity of the illumination. This means that a panel oriented using our approach will converge at, or close to, the direction of maximum irradiance, even when the irradiance function has multiple modes.

Many other visual sensors have also been proposed to iteratively tilt a solar panel toward the orientation that yields maximum irradiance. Shading-based sensors~\cite{wangOnchipSensorLight2013, wangCMOSSelfpoweredMonolithic2014, wangSelfpoweredSingleaxisMaximum2015, wangDesignImplementationSun2013, zogbiDesignConstructionSun1984, haywoodSolarCollectorDrive1978, tinaIntelligentSuntrackingSystem2013, queroLightSourcePosition2001} use a vertical wall to cast a shadow onto multiple photodetectors. A tracking system then orients the panel such that the shadow disappears, which occurs when the sun is directly above the panel on a clear day. Position-sensing diodes have also been used to find the direction of the sun~\cite{rothDesignConstructionSystem2004, chenAnalogueSunSensor2007}. While a panel using either of these methods can be oriented toward the sun on a clear day, it is not guaranteed to find the orientation that yields maximum irradiance in arbitrary lighting. Pineda and Arredondo~\cite{pinedaDesignImplementationSun2012} approach the problem of finding the best panel orientation in a variety of sky conditions by using a large number of detectors with smaller fields-of-view to sample the irradiance at different orientations. The panel is then oriented in the direction of the detector that produces the largest measurement. Rather than sample the irradiance function, we iteratively tilt a panel toward the direction of maximum irradiance using differential measurements from just four detectors.

\section{The Irradiance of a Solar Panel}

In this section, we derive an expression for the irradiance of a solar panel as a function of its orientation. 

\subsection{Total Irradiance from the Visible Hemisphere}
\label{sec:radiometry}

\begin{figure}[t]
    \centering
    \includegraphics[]{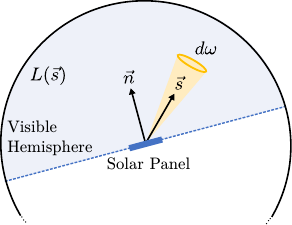}
    \caption{\textbf{Irradiance of a solar panel.} The illumination of a panel can be represented as a radiance function $\Ls$. The panel only receives light from the visible hemisphere, which is determined by the orientation $\nv$ of the panel. The total irradiance of the panel is computed as an integral of the radiance function over the visible hemisphere.}
    \label{fig:radiometry}
\end{figure}

The illumination seen by a solar panel can be completely described as a 4D light field~\cite{levoyLightFieldRendering1996}. Since the distances of light sources in any environment are typically much larger than the size of the panel, we can assume that the light incident from a specific direction is uniform over the panel's active area. Thus, as shown \cref{fig:radiometry}, the environmental illumination that the panel is exposed to can be represented as a 2D radiance function $\Ls$, where $\sv$ denotes the direction as seen from the center of the panel.

In the context of solar energy harvesting, we are specifically interested in the irradiance of the panel. Let $\nv$ be the vector normal to the panel. Then, the {\em directional irradiance} $\E$ from the direction $\sv$ and infinitesimal solid angle $\dw$ is
\begin{equation}
    \Esn = \Ls \, \max (\nv \cdot \sv, 0) \, \dw, \label{eq:Esn}
\end{equation}
where the $\max$ operator accounts for the fact that the panel only receives light from the hemisphere of illumination that is visible to the panel (see \cref{fig:radiometry}). 

We can now define the {\em total irradiance} $\ET$ of the panel as the integral of the directional irradiance (\cref{eq:Esn}) over the visible hemisphere,
\begin{equation}
    \ETn = \int_{\sv \in \mathbb{S}^2} \Ls \, \max (\nv \cdot \sv, 0) \, \dw, \label{eq:ET}
\end{equation}
where $\mathbb{S}^2$ is the set of points on the surface of the unit sphere.  $\ETn$ corresponds to the total power per unit area incident on the panel. {\em The goal of our work is to find, for any given environment, the panel orientation $\nv$ that maximizes the irradiance $\ETn$.}

\subsection{Irradiance as a Convolution on the Sphere}
\label{sec:ET-as-a-convolution}

From \cref{eq:ET}, we see that the irradiance function $\ETn$ can be written as the convolution of the radiance function $\Ln$ with a kernel defined as
\begin{equation}
    \kn = \max (\nv \cdot \hz, 0). \label{eq:k}
\end{equation}
Note that the kernel $\kn$ is centered about the zenith vector $\hz$. A detailed derivation that shows that the irradiance function is a convolution on the sphere~\cite{driscollComputingFourierTransforms1994} is given in \cref{app:convolution}. Notice that the kernel is simply the cosine of the zenith angle, but clipped to be non-negative. Given that the kernel is a smooth and broad function, it serves to simply low-pass filter the radiance function corresponding to the environment. An equivalent result in computer graphics shows that Lambertian reflectance low-pass filters the radiance function~\cite{basriLambertianReflectanceLinear2003, ramamoorthiRelationshipRadianceIrradiance2001}.

\begin{figure}[t]
    \centering
    \includegraphics[width=\linewidth]{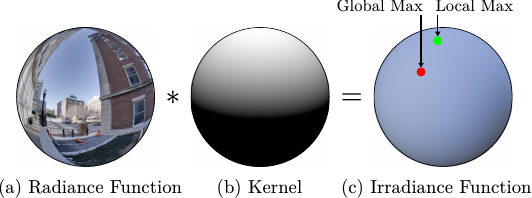}
    \caption{
    \textbf{Irradiance function as a convolution on the sphere.} The radiance function~(a) is convolved with the kernel~(b) to produce the irradiance function~(c). The irradiance function is a smooth version of the radiance function, but could still have multiple modes and hence multiple local maxima (red and green dots).
    }
    \label{fig:filtering-radiance-field}
\end{figure}

Since the irradiance function is the result of convolving the radiance function with a low-pass filter, it is guaranteed to be smooth. Consider the radiance function of a natural environment shown in \cref{fig:filtering-radiance-field}(a). Upon convolving it with the kernel in \cref{fig:filtering-radiance-field}(b), we get the irradiance function in \cref{fig:filtering-radiance-field}(c) which is very smooth. Despite its smoothness, however, the irradiance function may still have multiple modes, and hence multiple local maxima (red and green dots).

\section{Differential Light Sensing}

Imagine we had a method to measure the gradient of the irradiance function with respect to the solar panel orientation. Then, given any initial panel orientation, a naive approach would be to iteratively tilt the panel in the direction of the gradient and hope to converge at the direction of maximum irradiance. Unfortunately, this would not always work as we know that although the irradiance function is guaranteed to be smooth, it could have multiple local maxima (\cref{fig:filtering-radiance-field}(c)). In short, naive gradient ascent would likely get stuck at a local maximum, causing the panel to harvest less energy. In this section, we introduce our approach which uses differential light sensing to iteratively tilt a panel to arrive at the {\em global maximum} of the irradiance function, irrespective of the initial orientation of the panel or the complexity of the environmental illumination.

\subsection{The Photodifferential}
\begin{figure}[t]
    \centering
    \includegraphics[]{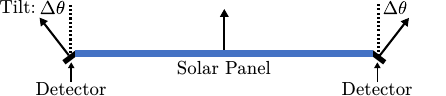}
    \caption{
    \textbf{Photodifferential sensor.} 
    Two tiny photodetectors on the sides of the solar panel are tilted with respect to the panel by a significant angle $\Dth$. The difference between the detector measurements is a finite difference of the irradiance function, which we refer to as the photodifferential.
    }
    \label{fig:differential-sensing}
\end{figure}

Consider a one-dimensional irradiance function $\ET(\theta)$, where the panel orientation is given by $\theta$. Our sensing method, illustrated in \cref{fig:differential-sensing}, produces a finite difference of the irradiance function $\ET(\theta)$ using just two photodetectors. The detectors are tilted by $\Dth$ in opposite directions with respect to the panel. Thus, the detectors measure the irradiance function at two different orientations: $\ET(\theta+\Dth)$ and $\ET(\theta-\Dth)$. The difference of these two measurements is proportional to a finite difference of the irradiance function. 
We refer to this difference as the ``photodifferential,'' which is defined as 
\begin{equation}
    \EDth = \frac{\ET(\theta+\Dth) - \ET(\theta-\Dth)}{2\Dth}.\label{eq:photodifferential}
\end{equation}

When the tilt angle $\Dth$ between the detectors and the panel is very small, the photodifferential is equal to the derivative (gradient) of the irradiance function with respect to the panel orientation. As stated earlier, this derivative is not of much use in our context---iteratively tilting the panel in the direction of the derivative will cause the panel  to get stuck at a local maximum. In the next section, we will show that the photodifferential produced by detectors with a \textit{large} tilt angle is actually the derivative of a blurred version of the irradiance function.  

\subsection{Detector Tilt Blurs the Irradiance Function}
We now analyze the effect of the detector tilt angle $\Dth$ on the photodifferential using Fourier analysis. Note that the photodifferential (\cref{eq:photodifferential}) can be written as the result of convolving the irradiance function $\ET(\theta)$ with the function 
\begin{equation}
    h(\theta) = \frac{1}{2\Dth} \left( \delta(\theta + \Dth) - \delta(\theta - \Dth) \right). \label{eq:photodiff-impulse}
\end{equation}
The Fourier transform of $h(\theta)$ is 
\begin{equation}
    \Hjw = \frac{1}{2\Dth} \left( e^{j\omega\Dth} - e^{-j\omega\Dth} \right). \label{eq:photodiff-freq-1}
\end{equation}
Using Euler's formula, $\Hjw$ can be written as
\begin{equation}
    \Hjw = \frac{j}{\Dth} \sin(\omega\Dth). \label{eq:photodiff-freq-2}
\end{equation}
By multiplying the numerator and denominator by $\omega$, \cref{eq:photodiff-freq-2} can be rewritten as
\begin{equation}
    \Hjw = \begin{cases}
        j\omega \ \frac{\sin(\omega\Dth)}{\omega\Dth} & \omega \ne 0\\
        0 & \omega = 0
    \end{cases}.\label{eq:photodiff-freq-3}
\end{equation}
Therefore, we get
\begin{equation}
    \Hjw = \underbrace{\textstyle j\omega}_{\mathclap{\text{\footnotesize{Derivative} \hspace{1mm}}}} \
    \underbrace{\textstyle \sinc(\omega\Dth)}_{\mathclap{\text{\hspace{1mm} \footnotesize{Box filter}}}}.\label{eq:photodiff-freq-final}
\end{equation}
\Cref{eq:photodiff-freq-final} shows that convolving the irradiance function with $h(\theta)$ is equivalent to first blurring the irradiance function with a box filter $\bth$, whose width is proportional to the detector tilt angle $\Dth$, and then taking the derivative.\footnote{Observe that $\lim_{\Dth\to0} \Hjw = j\omega$, which means $\Hjw$ becomes the continuous-time differentiator in the limit as the finite difference step size approaches 0. This aligns with the intuition that a finite difference computed using a very narrow box filter (infinitesimally small detector tilt) approximates the gradient of the irradiance function.}
That is,
\begin{equation}
    \EDth = \frac{d}{d\theta} \left( \Lth * \kth * \bth \right).\label{eq:photodiff-full}
\end{equation}
Dumont~\cite{dumontAnalysisFiniteDifference2015} presents a similar result for discrete signals showing that a finite difference with a large step size implicitly filters the signal with a moving average. This brings us to our first observation:
\begin{observation}
    The photodifferential $\EDth$ is the derivative of a blurred irradiance function, where the degree of blur is proportional to the detector tilt angle $\Dth$.
\end{observation}

\subsection{$E_D$ is the Derivative of a Unimodal Function}
\begin{figure}[t]
    \centering
    \includegraphics[]{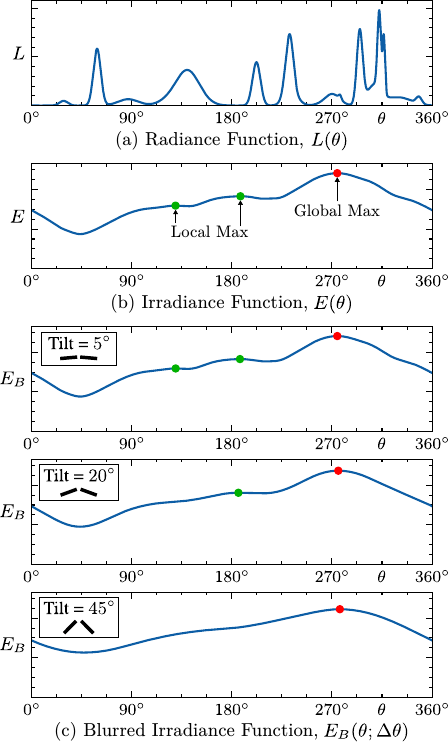}
    \caption{
    \textbf{A large detector tilt angle makes $\EB$ a unimodal function.}
    (a)~An example radiance function $\Lth$. 
    (b)~The irradiance function $\ETth$, which is the convolution of the radiance function with the kernel $\kth$, has multiple modes. 
    (c)~Increasing the detector tilt angle $\Dth$ has the effect of blurring the irradiance function, thereby eliminating local maxima. In this example, a tilt angle of $45\unit{\degree}$ results in a blurred irradiance function $\EB$ that is unimodal.
    }
    \label{fig:blurring}
\end{figure}

We now analyze the effect of the box filter $\bth$ in \cref{eq:photodiff-full}. Consider the one-dimensional radiance function $\Lth$ in \cref{fig:blurring}(a). Convolving it with the kernel $\kth$ produces the irradiance function $\ETth$ in \cref{fig:blurring}(b). This function has three modes: a global maximum (red dot) and two local maxima (green dots). In \cref{eq:photodiff-full}, we denote the function to which the derivative is being applied as the blurred irradiance function:  
\begin{equation}
    \EBth = \Lth * \kth * \bth.\label{eq:EB}
\end{equation}
In \cref{fig:blurring}(c), we plot $\EB$ for three different tilt angles: $5\unit{\degree}$, $20\unit{\degree}$, and $45\unit{\degree}$. We know that increasing the tilt angle has the effect of blurring the irradiance function with a wider box filter. Consequently, as the tilt angle increases, the number of local maxima in $\EB$ reduces. When the level of blur is sufficiently large ($\Dth=45\unit{\degree}$ for the example radiance function in \cref{fig:blurring}(a)), the local maxima disappear, rendering the blurred irradiance function unimodal. This brings us to our second observation:
\begin{observation}
    When the detector tilt angle $\Dth$ is sufficiently large, the blurring induced by the photodifferential creates a blurred irradiance function that is unimodal, regardless of the complexity of the illumination.
\end{observation}

\subsection{Gradient Ascent using the Photodifferential}

Since the photodifferential measures the derivative of a blurred irradiance function $\EB$ that has a single mode, our approach is to iteratively tilt the solar panel in the direction of the photodifferential to find the panel orientation that yields the {\em global maximum} of $\EB$. This approach, in effect, is equivalent to applying gradient ascent to $\EB$. 
However, we must ensure that the blurring from the detector tilt never introduces a new mode in $\EB$ that was not present in the original irradiance function.

\subsection{Can a Large Tilt Angle Introduce New Modes?}
The detector tilt angle must satisfy two conditions: (a)~The tilt angle should be large enough such that the blurred irradiance function $\EB$ is unimodal for virtually any environmental illumination. (b)~The tilt angle should not introduce new modes in the blurred irradiance function. This brings us to a well-known result in scale space theory: Blurring any function with a Gaussian cannot create new modes that were not present in the original function~\cite{koenderinkStructureImages1984, babaudUniquenessGaussianKernel1986, yuilleScalingTheoremsZero1986}. 

\begin{figure}[t]
    \centering
    \includegraphics[]{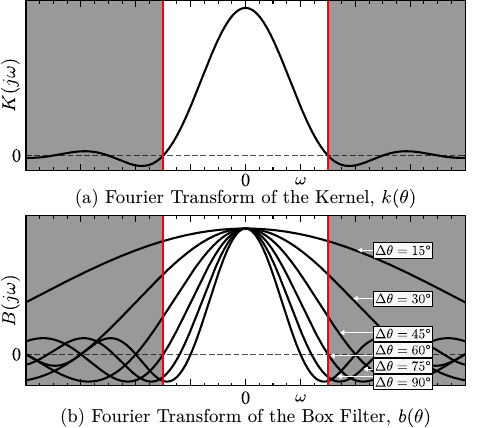}
    \caption{\textbf{The detector tilt angle has the effect of blurring the irradiance function with a Gaussian.} 
    (a)~The Fourier transform $\Kjw$ of the kernel $\kth$ is close to zero beyond its first zero crossings (red lines). This means that the irradiance function only has frequencies within the band between the red lines. 
    (b)~The Fourier transform $\Bjw$ of the box filter for a wide range of detector tilt angles. Observe that $\Bjw$ can be reasonably well-approximated by a Gaussian within the band between the red lines, even for the largest tilt angle of $\Dth=90\unit{\degree}$.
    }
    \label{fig:gaussian-approximation}
\end{figure}

Recall that the irradiance function $\ETth$ is the result of convolving the radiance function $\Lth$ with the kernel $\kth$. \Cref{fig:gaussian-approximation}(a) shows the Fourier transform $\Kjw$ of the kernel, which is close to zero for frequencies beyond the first zero-crossings, denoted by the two red lines. In other words, the kernel suppresses frequencies in the irradiance function that are outside the frequency band between the red lines. As a result, we are only concerned with frequencies of the irradiance function that lie within this band. 
\Cref{fig:gaussian-approximation}(b) shows the Fourier transform $\Bjw$ of the box filter $\bth$ for a wide range of detector tilt angles. Within the frequency band of the red lines, $\Bjw$ can be well-approximated as a Gaussian, even for the largest tilt angle of $\Dth=90\unit{\degree}$. This brings us to our third observation:
\begin{observation}
    Due to the bandwidth of the kernel $\kth$, large detector tilt angles do not introduce new modes in the blurred irradiance function.
\end{observation}

\section{Simulations}
\label{sec:simulations}

\begin{figure*}[t]
    \centering
    \includegraphics[width=\linewidth]{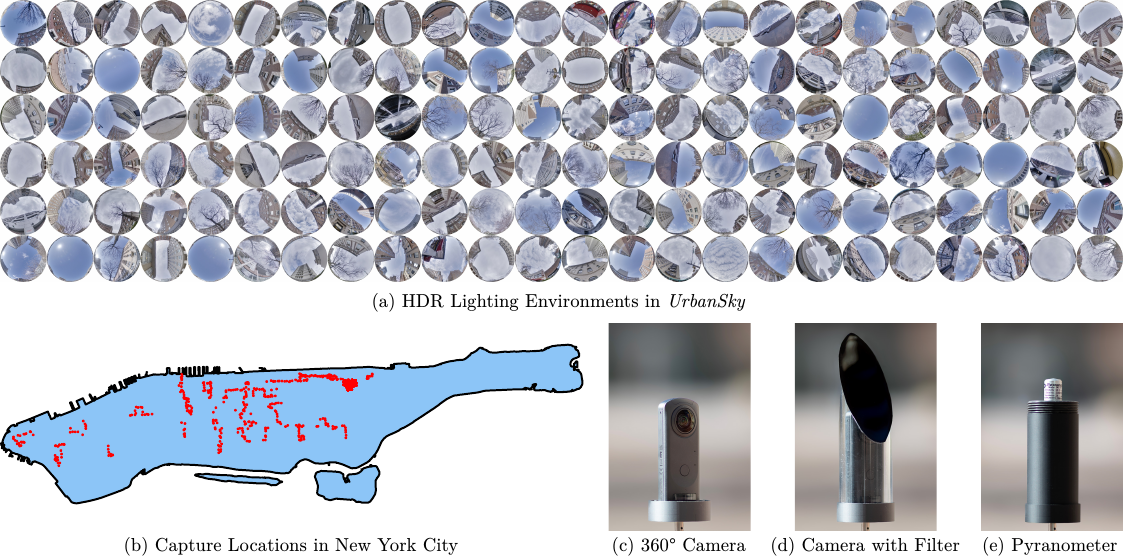}
    \caption{\textbf{\textit{UrbanSky}: Outdoor HDR Lighting Environments in New York City.}
    (a)~The dataset consists of 1,067 HDR lighting environments at different locations, times of day, and under varying weather conditions, a few of which are shown here. Each sphere shows the illumination at a particular location and time of day (tonemapped for visualization).
    (b)~\textit{UrbanSky} consists of lighting environments captured across New York City. Each red dot on the map indicates the location of a single capture. 
    (c)~We used a $360\unit{\degree}$ camera to capture the environmental illumination at each location. This camera captures a set of bracketed images that are merged and stitched into an HDR panorama.
    (d)~To faithfully capture the radiance from the sun without saturation, we place a neutral density filter between the camera and the sun to collect a second set of bracketed images. The neutral density filter is mounted on an acrylic pipe that can be easily slid over the camera without disturbing the camera's position.
    (e)~For each lighting environment, we also measure the global horizontal irradiance using a silicon pyranometer placed directly above the camera. 
    \textbf{Please see the \urbanskyweb{\textit{UrbanSky} website} to explore and download the dataset.}
    }
    \label{fig:urbansky}
    \vspace{-0.1in}
\end{figure*}

We have shown that large tilt angles are beneficial since they eliminate local maxima of the irradiance function and do not introduce new modes. In this section, we explore via simulations the range of detector tilt angles for which the irradiance of the panel (and hence the harvested energy) is at, or close to, the maximum. In addition, we conduct simulations to compare the energy harvested using our approach with alternative strategies for orienting a solar panel.

\subsection{\textit{UrbanSky}: A Dataset of Urban Lighting Environments}
\label{sec:urbansky}

To simulate our approach in real-world lighting environments, we have collected a large dataset of 1,067 HDR lighting environments across New York City. A few of the lighting environments in the dataset are shown in \cref{fig:urbansky}(a). \Cref{fig:urbansky}(b) shows the location of each captured environment overlaid on a map of New York City. At each location, we capture the environmental illumination using the $360\unit{\degree}$ camera (Ricoh Theta Z1) shown in \cref{fig:urbansky}(c). Since the dynamic range of outdoor scenes is very large, we capture $7\times$ bracketed images to create a single, HDR panorama. Unfortunately, on a sunny day, the sun's disk will be saturated in all of the captured images since the camera's exposure cannot be lowered beyond a certain point~\cite{stumpfelDirectHDRCapture2006}.
To faithfully measure the radiance from the sun, we capture a second set of $7\times$ bracketed images with an OD4 neutral density filter (Kodak Wratten 2) between the camera and the sun, as shown in \cref{fig:urbansky}(d). We then use PTGui Pro~\cite{PTGui} to merge and stitch the two sets of bracketed images into two $7400 \times 3700\,\unit{\text{px}^2}$ HDR panoramas.
Finally, we compute the radiance from the sun's disk using the panorama with the neutral density filter and overlay those values onto the panorama without the neutral density filter. This process ensures that none of the measurements of the radiance function are saturated. 

Alongside each captured environmental illumination, we also measure the global horizontal irradiance using a silicon pyranometer (EKO Instruments ML-01) placed directly on top of the $360\unit{\degree}$ camera (see \cref{fig:urbansky}(e)). For each environment in the dataset, we include metadata that specifies the global horizontal irradiance (in $\unit{W\per m\squared}$), the capture date, time of day, GPS location, and current weather conditions. In \cref{app:urbansky}, we provide additional details about the pyranometer, the image capture process, and post-processing steps.

\textit{UrbanSky} focuses specifically on capturing the complex illumination in dense urban settings caused by shadows, reflections, and changing weather conditions. To this end, we have chosen 49 specific locations on the campus of Columbia University at which we repeatedly capture the illumination at least 10 different times under varying weather conditions and at different times of day. We then used feature matching to automatically register the captured illumination at each location. In total, we have captured 526 lighting environments at specific locations at Columbia University, which comprise roughly half of \textit{UrbanSky}. The remaining half of the dataset is comprised of 541 environments captured across New York City (see \cref{fig:urbansky}(b)). The \urbanskyweb{\textit{UrbanSky} website} includes an interactive viewer to explore the data captured at each environment. While there are existing datasets of $360\unit{\degree}$ images of real environments~\cite{gardnerLearningPredictIndoor2017, hold-geoffroyDeepSkyModeling2019}, \textit{UrbanSky} focuses specifically on the illumination in urban environments. Furthermore, the pyranometer's measurement of global horizontal irradiance captures the total power per unit area that would be received by a silicon photovoltaic placed in the scene. We have released the entire dataset \urbanskyweb{online} to encourage future work both in solar energy and computer graphics.

\subsection{Simulations using \textit{UrbanSky}}
\label{sec:sim-urbansky}
We divided the radiance functions in \textit{UrbanSky} into two sets: one with 281 multimodal irradiance functions and the second with 786 unimodal irradiance functions. In our simulations, we assume the energy harvested by a panel to be proportional to its irradiance. For each of several tilt angles between $0\unit{\degree}$ and $90\unit{\degree}$, and for each radiance function within one of the two sets, we used gradient ascent to iteratively orient a panel starting from many initial orientations and then found the harvested energy as an average over the converged orientations. This energy is expressed as a percentage of the energy harvested using the optimal orientation (the global maximum). Once this is done for all the radiance functions in the set, and for each detector tilt angle, an average percentage harvested energy is computed. 

\begin{figure}[t]
    \centering
    \includegraphics[]{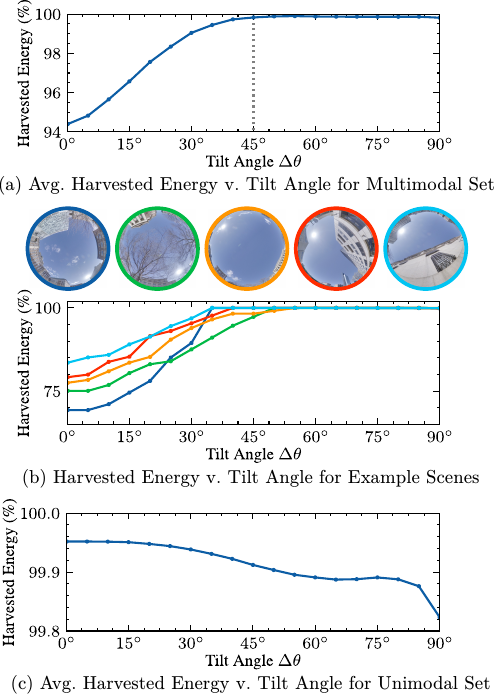}
    \vspace{-0.05in}
    \caption{
    \textbf{Simulations using real-world radiance functions.}
    (a)~The average energy harvested by a solar panel (as a percentage of the energy corresponding to the global maximum), plotted as a function of the detector tilt angle for the set of 281 multimodal irradiance functions. A large tilt angle of $\Dth=45\unit{\degree}$ (gray line) increases the harvested energy by $5.8\%$ on average compared to the smallest tilt angle. In general, any tilt angle of $\Dth=45\unit{\degree}$ up to $\Dth=90\unit{\degree}$ achieves a similar performance. 
    (b)~A few radiance functions that show that using a large detector tilt angle increases the harvested energy by up to $41\%$.
    (c)~The average harvested energy as a function of the detector tilt angle for the set of 786 unimodal irradiance functions. The harvested energy decreases ever so slightly with increasing detector tilt angles, which can be attributed to the global maximum shifting under increasing levels of blur. 
    }
    \label{fig:simulations}
\end{figure}

For the multimodal set, the average percentage energy harvested by the panel as a function of the detector tilt angle is shown in \cref{fig:simulations}(a). A tilt angle of $\Dth=45\unit{\degree}$ (denoted by the gray line) harvests $5.8\%$ more energy on average than the smallest tilt angle. In general, any tilt angle of $\Dth=45\unit{\degree}$ up to $90\unit{\degree}$ performs best, achieving a similar average harvested energy. \Cref{fig:simulations}(b) shows the percentage energy harvested as a function of the tilt angle for five radiance functions chosen from the multimodal set. As can be seen from the plots, the use of a large tilt angle results in a gain in harvested energy of up to $41\%$ in some scenes when compared with the smallest tilt. For the unimodal set, the average percentage harvested energy as a function of the detector tilt angle is shown in \cref{fig:simulations}(c). Note that the harvested energy decreases ever so slightly with increasing detector tilt angles. This can be attributed to the fact that the global maximum of the irradiance function shifts as the level of blur increases, causing the panel to converge close to but not exactly at the global maximum.

These simulations indicate that the use of a large tilt angle increases the harvested energy in a diverse set of complex lighting environments. In all subsequent experiments reported in the paper, we use a tilt angle of $\Dth=45\unit{\degree}$ since that angle is large enough to significantly increase the harvested energy in multimodal scenes, but not so large that it shifts the global maximum. In general, however, any tilt angle of $45\unit{\degree}$ or larger yields similar results.

\subsection{Comparison with Other Minimal Sensing Strategies}
Now that we have chosen a tilt angle, we use \textit{UrbanSky} to compare the energy harvested using our approach with other minimal sensing strategies for orienting a solar panel. For this comparison, we benchmark our approach against prior work that specifically focus on finding the best panel orientation in arbitrary lighting environments. We compare the energy harvested using our approach with three other minimal sensing strategies: (a) three detectors arranged on a tetrahedron~\cite{awayDualaxisSunTracker2017}, (b) two pairs of detectors separated by a shading wall~\cite{wangSelfpoweredSingleaxisMaximum2015}, and (c) forty detectors arranged on a geodesic dome~\cite{pinedaDesignImplementationSun2012}. The methods for orienting a solar panel using detectors on a tetrahedron~\cite{awayDualaxisSunTracker2017} or separated by a shading wall~\cite{wangSelfpoweredSingleaxisMaximum2015} are iterative methods that tilt the solar panel based on the difference of measurements until the orientation converges. The approach using forty detectors on a geodesic dome~\cite{pinedaDesignImplementationSun2012} orients the panel in a single step in the direction of the detector that produces the largest measurement. In addition, we compare the energy harvested using these approaches with the energy harvested by a solar panel pointing straight up without any visual sensing. We do not compare against methods for tracking the sun within a very narrow field-of-view (e.g.~using a position sensing diode~\cite{rothDesignConstructionSystem2004}), as these approaches are not intended to orient a solar panel starting from any initial orientation (instead, they are typically used to refine the orientation of a solar panel whose initial orientation is given by the apparent position of the sun in the sky).

The performance of each method for orienting a solar panel is summarized in \cref{tab:urbansky-benchmark}. For each approach, we report the average energy harvested, as a percentage of the energy that could be harvested at the global maximum, over all 1,067 scenes in \textit{UrbanSky}. Our approach using just four tilted detectors harvests the most energy, on average, and provides up to a $58\%$ increase in the average percent harvested energy compared to alternative methods. The code to benchmark all of these methods using \textit{UrbanSky} is available \projectweb{online}.

\subsection{Energy Harvested under Dynamic Illumination}
\label{sec:simulations-city}

\begin{figure*}[p]
    \centering
    \includegraphics[width=0.99\linewidth]{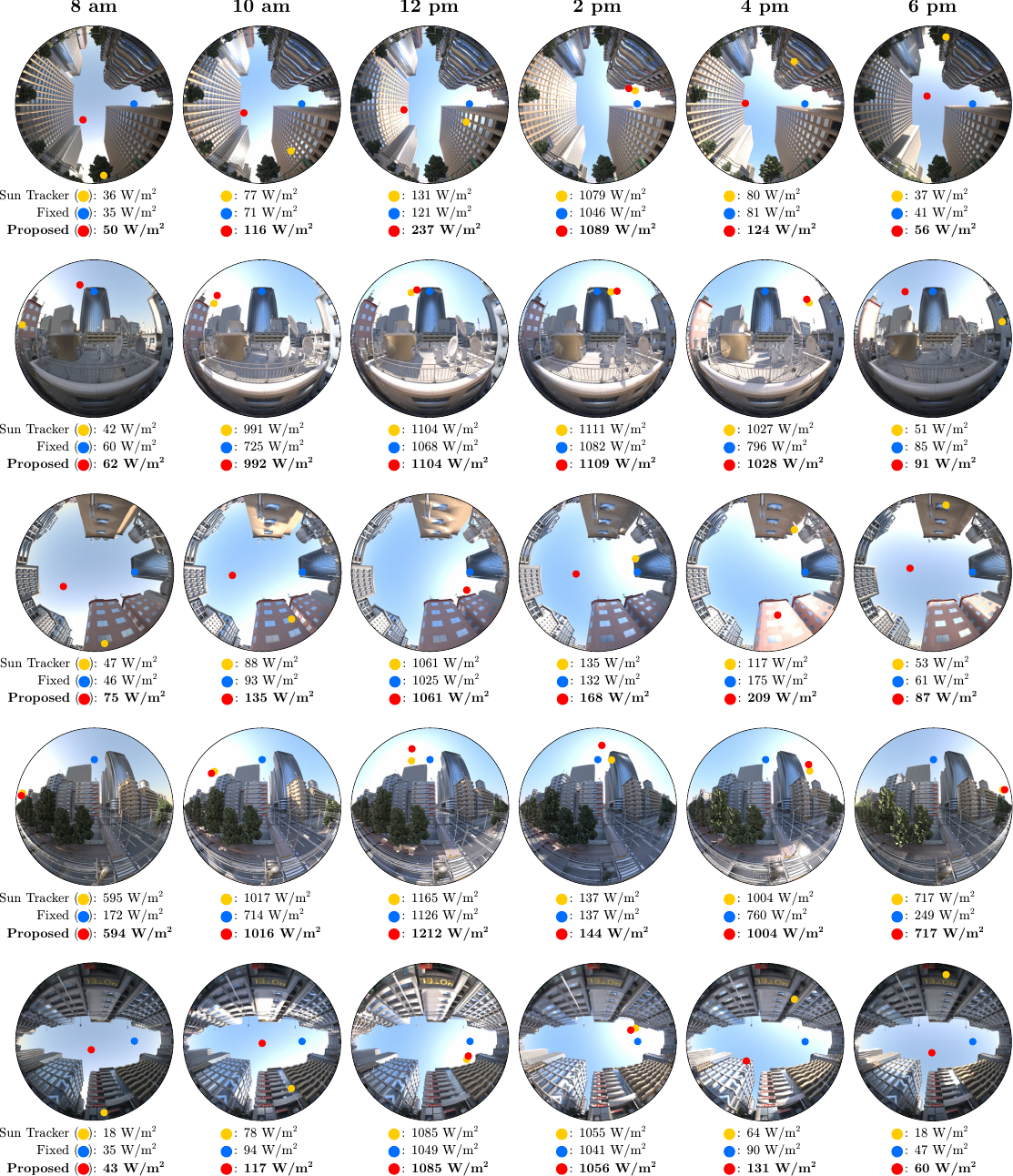}
    \vspace{-0.1in}
    \caption{
    \textbf{Simulations using dynamic illumination.}
    We used a physically based renderer to simulate the illumination of a solar panel in dense urban settings over a single day. We randomly chose 6,137 different solar panel positions that represent typical urban solar panel installations. Each row shows the illumination at a single location at different times of day. Notice that as the sun moves through the sky, the illumination of nearby buildings and objects can change dramatically. Every 10 minutes, we simulated the energy harvested by a solar panel oriented using our approach (Proposed). We compare its harvested energy with that of a panel that tracks the sun (Sun Tracker) and another in a fixed orientation (Fixed). In each scene, the orientations of the sun tracker, the fixed panel, and the proposed method are denoted using the yellow, blue, and red dots, respectively. We also include the irradiance of each solar panel below. Notice that across the diverse environmental illuminations shown here, the irradiance of the panel oriented using our approach (Proposed) is consistently higher than the irradiance of the other two panels. 
    }
    \label{fig:city-simulation}
\end{figure*}

So far, we have simulated our approach using real radiance functions captured in New York City at single instances in time. We now conduct simulations with dynamic illumination to demonstrate the gain in harvested energy over an entire day using our approach compared to alternative strategies for orienting a solar panel. 

To simulate dynamic illumination, we rendered physically accurate radiance functions that a solar panel would see throughout an entire day in an urban environment. Our simulation uses a physically based 3D model of a city, which includes complex geometry and physically accurate reflectance models. We selected 6,137 different solar panel locations by randomly sampling positions near the ground, on building walls, and on rooftops, which represent typical panel locations in a city.
Our goal is to simulate the illumination that would be seen by each of the 6,137 solar panels throughout the day. To this end, we used Mitsuba~\cite{wenzeljakobMitsuba3Renderer2022}, a physically based renderer, to generate the radiance function at each location every 10 minutes. To model the illumination from the sky at different times of day, we used the Ineichen and Perez model~\cite{ineichenNewAirmassIndependent2002a} to compute the radiance from the sun and the Preetham sky model~\cite{preethamPracticalAnalyticModel1999a} to compute the radiance from the sky. Each row in \cref{fig:city-simulation} shows rendered images at example locations at different times of day. Notice that as the sun moves through the sky, shadows move across the buildings, and some structures produce specular highlights. 

At each of the 6,137 locations, we simulate our approach for orienting a solar panel. Every 10 minutes at each location, we use the photodifferential to iteratively tilt the panel until convergence (the orientation converges when the magnitude of the photodifferential is smaller than a threshold). We then assume the panel irradiance to be constant for the entire 10-minute interval. Given the panel irradiance, we estimate the energy that would be harvested by a small (roughly $25\times25\,\unit{\cm\squared}$) solar panel with a typical efficiency of $20\%$. Furthermore, we also estimate the energy that would be consumed by the actuator.\footnote{The estimated energy consumed by the actuator is based on real measurements made using a solar panel of the same size.} At each location, we compute the net energy harvested over the entire day, after subtracting the energy consumed by the actuator.

\begin{table}[t]
    \caption{
    \textbf{Comparison with other minimal sensing strategies using \textit{UrbanSky}.}
    Over all 1,067 scenes in \textit{UrbanSky}, our approach for orienting a solar panel yields increases in harvested energy compared to alternative approaches (up to $58\%$ more harvested energy than alternative strategies). Included in this comparison is the energy harvested by a fixed solar panel pointing up, which does not use any visual sensing. 
    }
    \label{tab:urbansky-benchmark}
    \centering
    \vspace{0.1in}
    {\small
    {\setlength{\tabcolsep}{6pt}
    \begin{tabular}{m{2.9cm}cc}
        \toprule
        \multicolumn{1}{l}{Method} & 
        \multicolumn{1}{c}{\shortstack{Average\\Harvested Energy}} & 
        \multicolumn{1}{c}{\shortstack{Number of\\Sensors}} \\
        \midrule
        Tetrahedron~\cite{awayDualaxisSunTracker2017} & 63.4\% & 3 \\
        Shading Wall~\cite{wangSelfpoweredSingleaxisMaximum2015} & 84.8\% & 4 \\
        Fixed (Pointing Up) & 86.9\% & 0 \\
        Geodesic Dome~\cite{pinedaDesignImplementationSun2012} & 88.4\% & 40 \\
        \textbf{Ours} ($\Dth=45\unit{\degree}$) & \textbf{99.9\%} & 4 \\
        \bottomrule
    \end{tabular}
    }
    }
\end{table}

\subsection{Comparison with Alternative Strategies}

We compare the net energy harvested by a panel oriented using our approach with two panels oriented using common methods: one that tracks the sun (Sun Tracker) and another in a fixed orientation (Fixed). The fixed panel is oriented toward the equator, with the the zenith angle equal to the panel's latitude~\cite{duffieSolarEngineeringThermal2006}. In the images in \cref{fig:city-simulation}, the colored dots indicate the orientation of each of the three panels, and the irradiances of the panels are listed below. Notice that the irradiance of the panel oriented using our approach (Proposed) is consistently higher than the irradiance of the panels oriented using alternative strategies.

The energy harvested by the three panels over the 6,137 locations is summarized in \cref{tab:city-harvested-energy}. Over all the solar panel locations, our approach harvests $17.3\%$ more energy on average than the sun tracker and $39.7\%$ more energy on average than the fixed panel. The gain in harvested energy using our approach is due, in part, to the fact that the sun is often occluded by nearby buildings. In these cases, our approach orients the panel toward the sky and other well-illuminated buildings to increase the harvested energy (as an example, see the third image in the first row of \cref{fig:city-simulation}). 
Furthermore, our approach harvests, on average, $98.9\%$ of the energy that would by harvested by tracking the optimal orientation. Note that this is the net harvested energy after subtracting the energy consumed by the actuator. Before accounting for the actuator, our approach harvests $99.7\%$ of the energy harvested by tracking the optimal orientation. These results indicates that our approach consistently orients the panel toward the best orientation, and the small disparity in harvested energy is mostly due to the energy consumed by the actuator (which is $0.8\%$ of the total harvested energy, on average). Please refer to \cref{app:city} for the full distributions of the gain in harvested energy using our approach over the fixed panel and sun tracker, along with the distribution of the energy harvested using our approach as a fraction of the optimal orientation.

\begin{table}[t]
    \caption{
    \textbf{Performance comparison in dynamic illumination.}
    We compare the energy harvested by the sun tracker, the fixed panel, and the proposed method over 6,137 different locations in the simulated urban environment. Over a single day, the panel oriented using our method (Proposed) harvests more energy than the sun tracker and fixed panel and close to the energy that would be harvested by tracking the optimal orientation.
    }
    \label{tab:city-harvested-energy}
    \centering
    \vspace{0.1in}
    {\small
    \begin{tabular}{p{4cm}cc}
        \toprule
         & Mean & Median \\
        \midrule
        \textbf{Energy Gain Over} & & \\
        Sun Tracker & $17.3\%$ & $5.6\%$ \\
        Fixed & $39.7\%$ & $24.6\%$ \\
        \midrule
        \multicolumn{3}{l}{\textbf{Harvested Energy} {\footnotesize (\% of Optimal)}} \\
        Sun Tracker & $88.1\%$ & $93.8\%$ \\
        Fixed & $75.7\%$ & $79.7\%$ \\
        Proposed & $98.9\%$ & $99.3\%$ \\
        \bottomrule
    \end{tabular}
    }
\end{table}

\section{Experiments: Orienting a Solar Panel}

\begin{figure*}[p]
    \centering
    \includegraphics[]{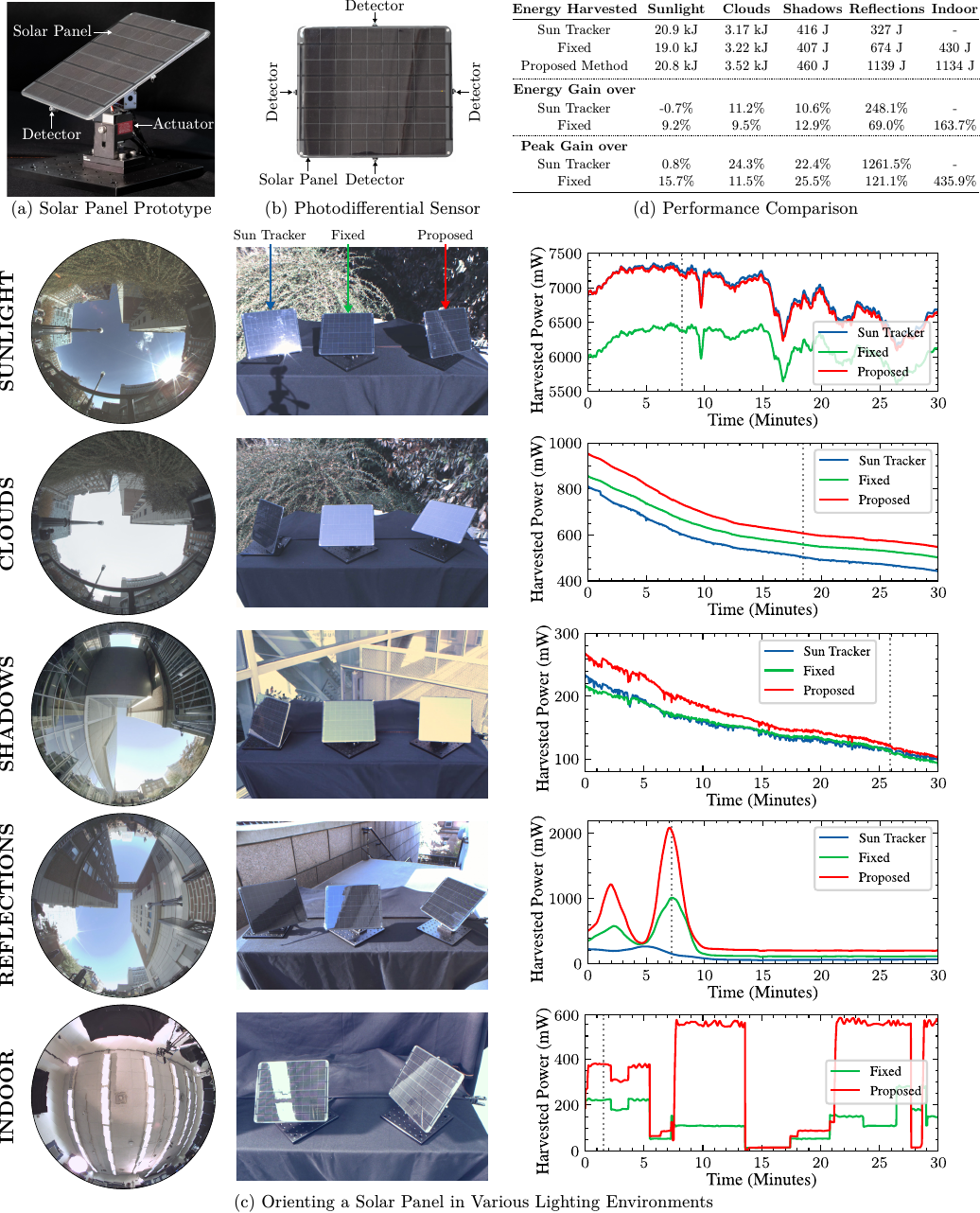}
    \vspace{-0.1in}
    \caption{
    \textbf{Experiments.}
    (a-b)~A prototype solar panel mounted on a two-axis actuator. A photodetector is attached to each of its sides with a tilt angle of $\Dth=45\unit{\degree}$.
    (c)~The performance of our panel (Proposed) is compared with panels using two widely used strategies: one that tracks the sun (Sun Tracker) and another in a fixed orientation (Fixed). Plots of the power harvested by the three methods are shown for five different scenarios: direct sunlight, cloudy sky, shadows cast by buildings, reflections from buildings, and an indoor workspace.
    The dotted line in each plot denotes the moment at which the fisheye image of the environment (left) and the image of the three panels (middle) were captured.
    (d)~The performances of the three panels are compared for the five scenarios in (c). Across a variety of real-world illumination phenomena, the proposed method increases the harvested energy.
    }
    \label{fig:experiments}
\end{figure*}

We developed the prototype shown in \cref{fig:experiments}(a), which includes a solar panel (Voltaic Systems P107C) mounted on a two-axis actuator ($2\times$ Miuezuth RDS3218). Four photodetectors (ams OSRAM BPW 34) are attached to the four sides of the panel with a tilt angle of $\Dth=45\unit{\degree}$, as seen in \cref{fig:experiments}(b). Since the illumination of outdoor environments generally varies slowly over the day, our prototype measures the photodifferential once every few minutes and then actuates the solar panel with a fixed step size of $5\unit{\degree}$ along both axes. To avoid oscillations in the panel orientation, the panel is only actuated when the magnitude of the photodifferential is above a threshold. Our prototype also includes a measurement system that continuously monitors the energy harvested by the solar panel at its maximum power point and the energy consumed by the actuator. \textit{The harvested energy reported in all of our experiments is the net harvested energy after subtracting the energy consumed by the actuator.} 

The goal of these experiments is to demonstrate a variety of real-world illumination phenomena in which our approach for orienting a solar panel increases the harvested energy compared to standard approaches. We refer the reader to \cref{sec:simulations} to evaluate the performance of our approach in a diverse set of real and simulated lighting environments.

In \cref{fig:experiments}(c), we show results for five different scenarios. In each case, we compare the harvested energy using our approach (Proposed) with a solar panel that tracks the sun (Sun Tracker) and another in a fixed orientation (Fixed). The trajectory of the sun tracker is determined by the position of the sun in the sky, which is computed using the location (latitude and longitude) of the panel and the date and time of day. The fixed panel is oriented toward the equator, with the angle from the zenith equal to the panel's latitude~\cite{duffieSolarEngineeringThermal2006}. The plot in the first row of \cref{fig:experiments}(c) shows the harvested power by each panel as a function of time in direct sunlight. The fisheye image shown on the left corresponds to the time denoted by the dashed line in the harvested power plot. As expected, the panel oriented using our method tracks the sun, and thus it harvests the same amount of energy as the sun tracker and $9.2\%$ more energy than the fixed panel over a one-hour period (see the table in \cref{fig:experiments}(d)). A video of the environmental illumination and moving panels is included on the \projectweb{project website}.

On a cloudy day, orienting a solar panel toward the sun is almost certainly not the optimal orientation. In the second row of \cref{fig:experiments}(c), the panels are placed outside on a cloudy day. Since our panel points up toward the clouds, it harvested $11.2\%$ more energy than the sun tracker and $9.5\%$ more energy than the fixed panel over a one-hour period. 

When a solar panel is deployed in a dense urban environment, the panel's view of the sun is often obstructed by nearby buildings. This case is shown in the third row of \cref{fig:experiments}(c), where the panels are in the shadow cast by nearby buildings. The panel oriented using our approach, however, orients itself toward the unobstructed portion of the sky. In this example, over a one-hour period, our panel harvested $10.6\%$ more energy than the sun tracker and $12.9\%$ more energy than the fixed panel (see \cref{fig:experiments}(d)). 

In urban settings, reflections of sunlight by nearby buildings can have a significant impact on the energy harvested by a panel. In the fourth row of \cref{fig:experiments}(c), the panels are placed in a shaded area that is briefly illuminated by strong reflections from a nearby building. The peaks in the plot of the harvested power correspond to times when the reflections were strong. At such times, our panel orients itself in the direction of the reflections. At the same time, the sun tracker points toward the wall of another building, as its orientation is determined by the position of the sun in the sky, regardless of nearby buildings that obstruct the panel's view of the sun. As reported in \cref{fig:experiments}(d), in this setting, our panel harvested $248.1\%$ more energy than the sun tracker and $69.0\%$ more energy than the fixed panel over a one-hour period. During the strongest reflection, the instantaneous harvested power in our case is more than $10\times$ that of the sun tracker (bottom two rows of \cref{fig:experiments}(d)). Please refer to the \projectweb{project website} for a video of the environmental illumination and solar panels.

In our final example (the fifth row of \cref{fig:experiments}(c)), the panels are placed in an indoor environment. Throughout the day, people working in the space turn on and off different floor lamps. Unlike the outdoors, in indoor spaces, the lighting can vary dramatically over time. Hence, in this case, we programmed our panel to iteratively adjust its tilt until convergence every minute. Compared to a fixed panel oriented toward the center of the ceiling, our method harvested $163.7\%$ more energy over a one-hour period. Please see the \projectweb{project website} for a video of the indoor scene.

\section{Discussion}
We have presented a minimal sensing method for iteratively orienting a solar panel to maximize its irradiance, thereby maximizing the energy harvested by it.
There are two directions we plan to pursue as future work.
First, we plan to deploy our method at a large scale by working with companies that install medium- and small-sized panels in cities and indoor environments. Second, we plan to explore the case in which it is preferable not to use any moving parts (actuators). In this case, we wish to find the optimal fixed orientation of a solar panel in any lighting environment while accounting for the complex, time-varying illumination seen from a particular location. This problem is particularly interesting for locations at which the scene's 3D structure and material properties are unknown. The optimization would need to account for nearby infrastructure, the variation of the sun's trajectory, and the historical weather conditions recorded for the location.

\section*{Acknowledgements}
This work was supported by the Office of Naval Research (ONR) awards N00014-23-1-2096 and N00014-21-1-2378.  Jeremy Klotz was supported by a National Defense Science and Engineering Graduate (NDSEG) Fellowship. The authors are grateful to Behzad Kamgar-Parsi at ONR for his support and encouragement. The authors also thank Mikhail Fridberg for his input on the prototype, Changxi Zheng for his feedback on the simulations, Bill Miller for his help with fabrication, Makoto Odamaki for technical discussions about the Ricoh Theta camera, and Jean-François Lalonde for sharing the Laval Indoor and Outdoor HDR datasets. Finally, the authors thank Joanne Chan for her help with collecting data for \textit{UrbanSky} and Lulu Wang for her help in creating the \textit{UrbanSky} website.

\appendix
\gdef\thesection{\Alph{section}} 
\makeatletter
\renewcommand\@seccntformat[1]{\appendixname \csname the#1\endcsname.\hspace{0.5em}}
\makeatother

\section{Irradiance as a Convolution}
\label{app:convolution}
The irradiance function $\ETn$ is defined in \cref{eq:ET} as
\begin{equation}
    \ETn = \int_{\sv \in \mathbb{S}^2} \Ls \, \max (\nv \cdot \sv, 0) \, \dw. \label{eq:ET-supp}
\end{equation}
Here we show that $\ETn$ can be written as the convolution of the radiance function $\Ln$ with a kernel $\kn$, which is defined in \cref{eq:k} as
\begin{equation}
    \kn = \max (\nv \cdot \hz, 0), \label{eq:k-supp}
\end{equation}
where $\hz$ is the zenith vector.

When convolving a function defined on the surface of the sphere with a kernel, the kernel is first rotated and then multiplied with the original function. Let $R_{\sv}$ be a rotation operator that rotates the zenith vector $\hz$ to $\sv$, i.e.~$\sv = R_{\sv}\hz$. \Cref{eq:ET-supp} can now be written as
\begin{equation}
    \ETn = \int_{\sv \in \mathbb{S}^2} \Ls \, \max (\nv \cdot R_{\sv}\hz, 0) \, \dw. \label{eq:ET-2-supp}
\end{equation}
Using the fact that $\nv \cdot \Rs\hz = \Rs^{-1} \nv \cdot \hz$, we can rewrite \cref{eq:ET-2-supp} as
\begin{equation}
    \ETn = \int_{\sv \in \mathbb{S}^2} \Ls \, \max (\Rs^{-1} \nv \cdot \hz, 0) \, \dw. \label{eq:ET-3-supp}
\end{equation}
Substituting \cref{eq:k-supp} into \cref{eq:ET-3-supp}, $\ETn$ becomes
\begin{equation}
    \ETn = \int_{\sv \in \mathbb{S}^2} \Ls \, k(\Rs^{-1} \nv) \, \dw. \label{eq:ET-4-supp}
\end{equation}
\Cref{eq:ET-4-supp} shows that the irradiance function $\ETn$ is the convolution of the radiance function $\Ln$ with the kernel $\kn$.\footnote{Strictly speaking, the convolution on a sphere is defined by integrating over the three degrees of freedom of the rotation operator~\cite{driscollComputingFourierTransforms1994}. Even though \cref{eq:ET-4-supp} only integrates over two degrees of freedom, adding the third degree of freedom does not change the result in our context. Since the kernel $\kn$ is symmetric about the $z$-axis, $\kn$ is invariant to rotations about the $z$-axis. Thus, integrating over the rotation operator's third Euler angle, $\psi$, only scales the expression by a constant, leaving the convolutional form unchanged.}
Both Ramamoorthi and Hanrahan~\cite{ramamoorthiRelationshipRadianceIrradiance2001} and Basri and Jacobs~\cite{basriLambertianReflectanceLinear2003} present a similar derivation showing that Lambertian reflectance has the effect of convolving the radiance function with the kernel defined in \cref{eq:k-supp}. 
We refer the interested reader to Driscoll and Healy~\cite{driscollComputingFourierTransforms1994} for a detailed introduction to the convolution of two functions defined on the surface of the sphere.

\section{Details of the \textit{UrbanSky} Dataset}
\label{app:urbansky}

For each scene in \textit{UrbanSky}, we use a silicon pyranometer (EKO Instruments ML-01) to measure the global horizontal irradiance. We have chosen the pyranometer such that its spectral response and directional response closely match that of a silicon photovoltaic~\cite{ekoinstrumentsML01SiliconPyranometer}. Thus, the irradiance measurement corresponds to the power per unit area that would be received by a silicon photovoltaic pointing up in the scene. During the capture process, the neutral density filter in \cref{fig:urbansky}(d) is only needed to capture the radiance from the sun's disk. On cloudy days, this filter is not needed since the camera can capture a single set of $7\times$ bracketed images without any saturation. Thus, we only use the neutral density filter on sunny days. Finally, as a post-processing step, we used EgoBlur~\cite{rainaEgoBlurResponsibleInnovation2023} to automatically detect and blur faces and license plates in the captured images. We then manually reviewed each image to blur faces and license plates that were not automatically detected by EgoBlur.

\section{Details of Simulation Results}
\label{app:city}
\begin{figure}[t]
    \includegraphics[]{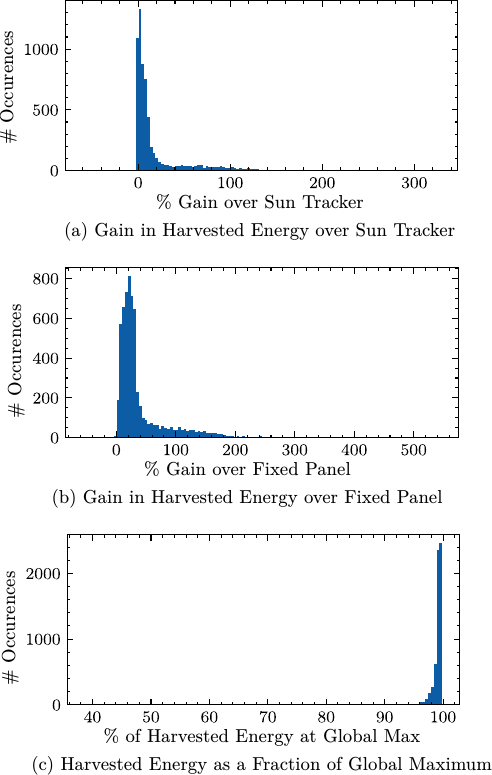}
    \centering
    \caption{
    \textbf{Energy harvested by solar panels in simulated dynamic illumination.}
    (a)~Histogram of the percent gain in net energy harvested by the panel oriented using our approach compared to the sun tracker.
    (b)~Histogram of the percent gain in net energy harvested using our approach compared to the fixed panel.
    (c)~Histogram of the net energy harvested using our approach compared to the energy that could be harvested by tracking the optimal orientation. All histograms show the distribution of results over 6,137 solar panel locations.
    }
    \label{fig:app-city-distributions}

\end{figure}

Here we include histograms of the results from simulating the energy harvested by solar panels at 6,137 different locations in an urban environment (as described in \cref{sec:simulations-city}). \Cref{fig:app-city-distributions}(a,b) show the histograms of the percentage gain in net energy harvested over the entire day using our approach compared to the sun tracker and fixed panel. The statistics of these distributions are summarized in the first two rows of \cref{tab:city-harvested-energy}. \Cref{fig:app-city-distributions}(c) shows the histogram of the percentage of energy harvested over the entire day using our proposed method compared to the energy that could be harvested using the optimal orientation, which corresponds to the last row of \cref{tab:city-harvested-energy}. As explained in in \cref{sec:simulations-city}, the harvested energy for each panel is the net harvested energy after subtracted the energy consumed by the actuator.

\bibliographystyle{elsarticle-num} 
\bibliography{Solar.bib}

\end{document}